\documentclass{article}
\usepackage{natbib}
\usepackage{graphicx} 
\usepackage{amsmath,amssymb}
\usepackage{setspace}
\usepackage{subcaption}
\usepackage{tikz}
\usepackage{url}
\usetikzlibrary{cd}

\newcommand{\cpt}[1]{\texttt{#1}}

\title{What Machine Learning Tells Us About the Mathematical Structure of Concepts}
\author{
  Jun Otsuka$^{1,2,3}$\thanks{\texttt{junotk@gmail.com}} \\
  \footnotesize $^1$The Department of Philosophy, Kyoto University \\
  \footnotesize $^2$Data Science and AI Innovation Research Promotion Center, Shiga University \\
  \footnotesize $^3$Center for Advanced Intelligence Project, RIKEN
}

\date{\today}

\begin{document}

\maketitle

\begin{abstract}
    This paper examines the connections among various approaches to understanding concepts in philosophy, cognitive science, and machine learning, with a particular focus on their mathematical nature.
    By categorizing these approaches into Abstractionism, the Similarity Approach, the Functional Approach, and the Invariance Approach, the study highlights how each framework provides a distinct mathematical perspective for modeling concepts. The synthesis of these approaches bridges philosophical theories and contemporary machine learning models, providing a comprehensive framework for future research. This work emphasizes the importance of interdisciplinary dialogue, aiming to enrich our understanding of the complex relationship between human cognition and artificial intelligence.
\end{abstract}

\section{Introduction}
What are concepts? This is arguably one of the fundamental questions in philosophy, with nearly as many answers as there are philosophers.
Aristotle's \emph{Categories}, Locke's compositional theory of ideas, Kant's transcendental logic, Hegel's concrete universals, Cassirer's \emph{Funktionsbegriff}, and Wittgenstein's discussion of family resemblance are  just a few notable examples of philosophical theories on concepts.
Since the last century, philosophical inquiries into concepts have been further accelerated and enriched through interaction with cognitive science, which added an empirical dimension to the discussion. 
Various models of concepts, such as prototype theory, exemplar theory, and the so-called \emph{theory-theory}, have been proposed, oftentimes in opposition to or drawing inspiration from existing philosophical theories, and brought to empirical scrutiny \citep{Margolis1999-ob, Murphy2004-nk}. 
At the same time, the discussions highlighted the multifaceted nature of what we call concepts, prompting skepticism about the general theory of concepts that would unify all the empirical findings and the theoretical roles we expect for ``concepts'' \citep{Machery2009-nw}. 

On the other hand, the nature of concepts has also been a main topic in the machine learning and artificial intelligence literature since the first generation of AI scholars set out to seek appropriate knowledge representation to build expert systems \citep[cf.][]{McCarthy1969-ng}.
Today, the main impetus of machine learning has shifted from rule-based algorithms to statistical inference powered by deep neural networks (DNNs), which spurs an interest on how these models are extracting appropriate \emph{representations} from data and using them in various tasks including object classifications and texts or image generations, to name a few \citep[e.g.][]{Buckner2019-it, Buckner2023-wx}.
The swift advancement in AI research has developed hand in hand with the mathematical modeling of representations \citep[cf.,][]{Goodfellow2016-xx}.
The ``Good-Fashioned Old AI'' or GOFAI employed a lattice-like structure to represent the knowledge base of expert systems. 
Contemporary natural language processing (NLP) and object recognition embed words and objects in some metric space (e.g. vector space). 
Generative models (such as variational auto-encoder or VAE; \citealt{Kingma2013-eo}) are thought to map data onto manifolds (intuitively, curved hypersurfaces) in the latent space. 
Recent studies use group theory to formulate the invariant and compositional features of concepts  \citep{Cohen2016-yq, Higgins2018-hm, Wang2024-wy, Weiler2024-ym}. 
These mathematical approaches aim to explicate the formal nature of machine representations from different aspects, with a view toward unlocking insights into the remarkable capabilities of DNNs. 

The aim of this paper is to put together insights from these different studies on concepts---in philosophy, cognitive science, and machine learning---in one place and draw connections among them. 
In particular, the paper will categorize these works into the following four types and characterize each type in terms of the corresponding theories of different traditions:
\begin{enumerate}
    \item \textbf{The Abstractionism}: is a classical Aristotelian view that concepts are formed through abstraction. This is the basis of the classical theory in cognitive science and formal ontology in expert systems, and is suitably represented by lattice algebra. 
    \item \textbf{The Similarity Approach}: underlies the Wittgensteinian theory of family resemblance (philosophy), prototype and exemplar theories (cognitive theory), and vector space representations in machine learning (NLP, object recognition). 
    \item \textbf{The Functional Approach}:  dates back to Lotze and Cassirer's \emph{funktionsbegriff}, and captures some aspects of the theory theory or knowledge-based approaches in cognitive science. A similar idea underlies various kinds of manifold learning and generative models (such as VAE), where representations are analyzed in terms of manifolds. 
    \item \textbf{The Invariance Approach}: draws on the idea that concepts must be invariant under certain groups of transformations \citep{Cassirer1944-pm, Jantzen2015-gx}. The idea is featured in the recent research on invariant or disentangled representations in machine learning (e.g., convolutional networks, $\beta$-VAE, and so on), and is formulated using the mathematical theory of groups \citep{Cohen2016-yq, Higgins2018-hm, Weiler2024-ym}.  
\end{enumerate}


Through this (admittedly partial and incomplete) classification, the present paper aims to reveal hitherto unrecognized commonalities among philosophical, psychological, and computational theories. 
This will facilitate researchers working in different disciplines to learn from each other and extrapolate findings in one domain to another. 
For instance, modeling practices in cognitive science and machine learning suggest how \emph{a priori} philosophical ideas may be implemented empirically and computationally. 
In turn, philosophical considerations may be instrumental in both exploring and refining scientific models, by setting desiderata or providing their interpretations.
Such interdisciplinary exchange is becoming increasingly important in the context of the recent explosive advancements in AI, which raises the critical question of how we can comprehend the thought processes of machines \citep{Otsuka2022-eq}. Since representations are expected to play key roles in both the generalizability and interpretability of deep learning models, explicating their structures and comparing them with human concepts should provide a deeper understanding of the mechanisms underlying AI and contribute to the development of more transparent and robust models.


With that said, this paper does not purport to make any empirical claim about the psychological mechanism underlying human reasoning. 
It is not the goal of this paper to identify the conceptual model that best captures how we humans think, for such a question should be answered not by philosophy but by empirical studies.  
Nor is the comparison meant to imply the similarity between machine and human. I remain completely agnostic about whether human brains \emph{are really like} neural networks (or vice versa). 
The focus of this paper is not on empirical implementations, but rather exclusively on models themselves. That is, its sole aim is to compare and analyze the (mathematical) nature of various theories of concepts, regardless of their empirical veracity. In this sense, the present paper is meant to serve as a prolegomenon for empirical studies, by taking stock and pulling together different approaches to modeling concepts. 

This paper unfolds as follows: Section 2 begins with an exploration of the classical Aristotelian theory of Abstractionism and its modern adaptations in cognitive science, expert systems, and lattice theory. Section 3 discusses the Similarity Approach, tracing its origins to Wittgenstein’s family resemblance and its application in prototype and exemplar theories within cognitive science, as well as in vector space models in machine learning. Section 4 examines the Functional Approach, which draws on Lotze and Cassirer’s work, and its parallels in manifold learning and generative models like VAEs. Section 5 introduces the Invariance Approach, highlighting the role of group theory in understanding invariant and equivariant representations in both philosophy and contemporary machine learning, particularly in the context of disentangled representations. Finally, Section 6 provides a comparative analysis of these approaches, discussing their implications for future research in the interdisciplinary study of concepts and representations.

\section{The Abstractionism}
According to abstractionism, concepts are formed through abstractions from individual data.
For instance, the concept \cpt{human} is formed from individual humans by abstracting away idiosyncratic differences, say in height, hair color, or any physical and psychological characteristics. 
The abstraction process can be done in stages. 
The concept \cpt{mammal} may be formed by abstracting away differences among \cpt{human}, \cpt{tiger}, \cpt{dog}, etc., and by repeating a similar process one can obtain more general concepts such as \cpt{animal}, \cpt{organism}, and so on. 
The resulting upside-down tree-like structure is called the Porphyrian tree. 
Going up and down through this tree respectively correspond to abstraction and specification. 
Each downward branching of the tree represents logical disjunction, such as \cpt{mammal} = \cpt{human} $\vee$ \cpt{tiger} $\vee$ ... 
Disjunction amounts to ignoring differences among the terms. 
Conversely, one may also create a concept through conjunction, like \cpt{centaur} = \cpt{human} $\wedge$ \cpt{horse}, whereby the concept of a centaur is formed by combining the features possessed by humans and horses.
Conjunction is represented by upward branching, which yields a lattice-like structure. 

This way of looking at concepts, which \cite{Heis2007-ti} dubbed as ``Aristotelian abstractionism,'' dates back to Aristotle's \emph{Categories} and reached a certain level of perfection in the modern period, as witnessed in Arnauld and Nicole's \emph{Port Royal Logic} and Kant's \emph{Logik} \citep{Igarashi2023-zh}.
Modern logic made an explicit distinction between \emph{extents} and \emph{intents}.
The extent is a set of objects to which a given concept applies, while its intent is a set of properties that defines the concept. 
For instance, the extent of \cpt{human} includes Socrates, Caesar, etc., while its intent would include such properties as being bipedal, rational, etc. 
A concept is fully identified by its intent, i.e., conditions that are separately necessary and jointly sufficient for being its instance, or in other words, the properties shared by all and only members of the corresponding extent. 

The notion reminiscent of abstractionism extends beyond the boundaries of philosophy. 
Because concepts are essential building blocks of human thought and understanding, cognitive scientists have developed several models of concepts to characterize their cognitive functionalities in perception, memory, reasoning, language processing, and decision-making. 
The \emph{Classical Theory} is one such theory that has often served as a foil for other views, and identifies a concept with the corpus of information pertaining to its defining attributes \citep{Margolis1999-ob, Murphy2004-nk, Machery2009-nw}.
Along with its name, the resemblance to abstractionism should be clear. 

The abstractionist conceptualization of concepts has also laid the foundational framework for initial AI research, notably influencing the development of the knowledge base for expert systems \citep[e.g.,][]{Sowa2000-ps}.
Expert systems aim to emulate the decision-making ability of human experts (e.g. medical doctors) by applying inferential rules to information stored in the knowledge base. 
The standard form of knowledge representation, often called \emph{ontology}, stores information obtained from experts in a hierarchical structure: for instance, \cpt{blood vessels} consist of \cpt{veins}, \cpt{capillaries}, and \cpt{arteries}, which are further broken down by size and location \citep{Lucas1991-gv}. 
In this way, ontologies in medical informatics, such as SNOMED CT\footnote{SNOMED CT: Systematized Nomenclature of Medicine-Clinical Terms, International Health Terminology Standards Development Organisation (IHTSDO). https://www.snomed.org/}, use abstractionist structures to organize concepts related to anatomy, diseases, and treatments, allowing for efficient retrieval and inference across different levels of generality. This enables expert systems to not only store and retrieve relevant knowledge but also to reason about it by navigating the hierarchical relationships between concepts.

\begin{figure}
    \centering
    \includegraphics[width=0.4\linewidth]{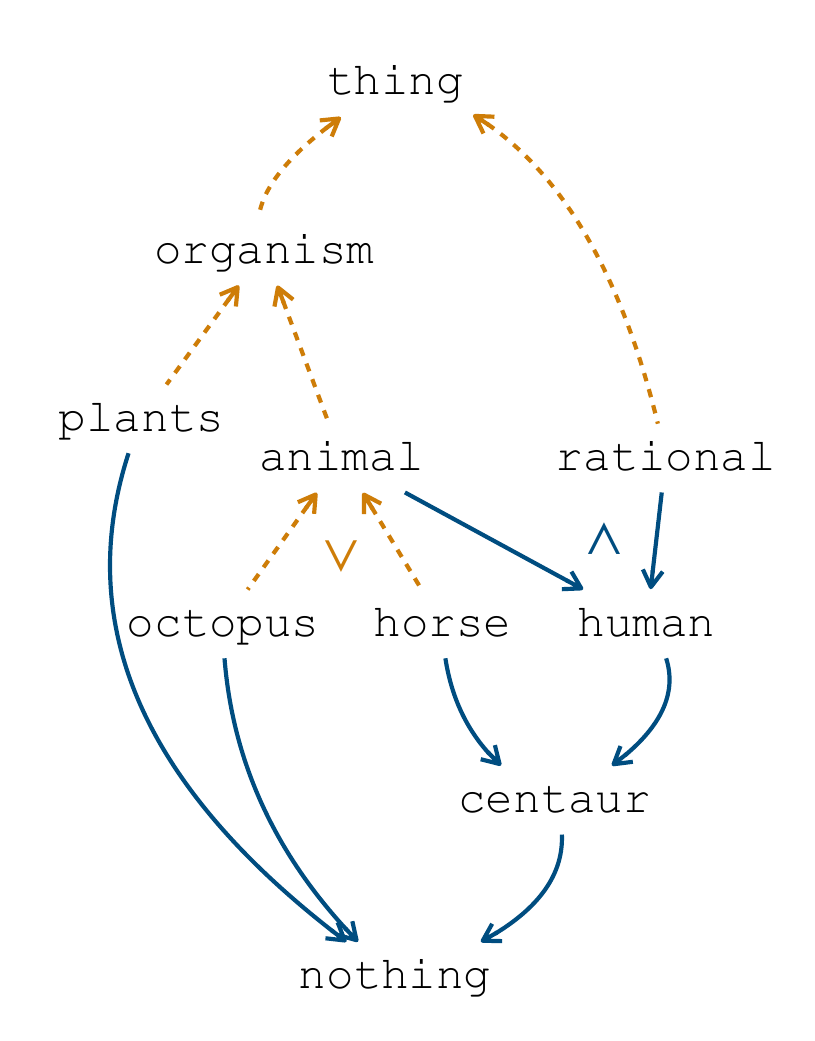}
    \caption{An example of a conceptual lattice. Each link represents `is-a' relationship between the relata. The upward arrows (dotted and orange) indicate the join operation that abstracts various concepts, while the downward arrows (solid and blue) indicate the meet operation that combines concepts. }
    \label{fig:lattice}
\end{figure}

The abstractionist hierarchy has a clear algebraic structure \citep{Heis2007-ti, Igarashi2023-zh}. 
The abstraction relationship such as ``$x$ is $y$'' (as in ``human is an animal'') is reflexive, antisymetric, and transitive, and thus defines a partial order $x \prec y$ over a set of concepts. 
Moreover, for a set of concepts $\{x_1, x_2, \dots, x_n\}$, one can define their join $\bigvee_i^n x_i$ as the concept formed by abstracting away their differences. 
Likewise, meet $\bigwedge_i^n x_i$ is the (most abstract) concept that belongs to all $x_i$'s. 
The whole conceptual hierarchy, therefore, can be represented by a (complete) \emph{lattice}.
The most abstract concept, such as \cpt{substance} or \cpt{thing} in the traditional Porphyrian tree, is the top element 1 of the lattice, while the bottom 0 would signifies the opposite concept of \cpt{nothing} (Fig. \ref{fig:lattice}). 

The algebraic structure of conceptual lattices has been studied in \emph{formal concept analysis} or FCA \citep[e.g.][]{Davey2002-ao}. 
FCA defines a concept $c$ as a pair $(I_c, E_c)$ of sets of its extent and intent, where $I_c$ contains just those properties shared by all the items in $E_c$, and similarly $E_c$ are precisely those items that share all the properties of $I_c$, so that $I_c$ gives the necessary and sufficient conditions for the membership of $E_c$. 
Then the families of intents $\{I_c\}_c$ and extent $\{I_c\}_c$ respectively form complete lattices of subsets of $I$ and $B$, where $I := \bigcup_c \{I_c\}_c$ and $E := \bigcup_c \{E_c\}_c$, i.e., all properties and items, respectively. 
These two lattices provide two different ways of looking at the abstraction relationship. 
On the intent side, $c'$ is more abstract than $c$ (i.e. $c \prec c'$) iff $I_{c'} \subset I_c$, where the difference $I_c \setminus I_{c'}$ is those properties that are ``abstracted away.''
On the extent side, $c \prec c'$ means $E_c \subset E_{c'}$, so that a more abstract concept has a wider extent. 
These two lattices are mirror images of each other: i.e., $c \prec c'$ iff $I_{c'} \subset I_c$ iff $E_c \subset E_{c'}$.
Thus the above simple definition neatly represents the hierarchical structure of concepts in two different but essentially equivalent ways, namely as the dual lattices of  intents and extents.\footnote{This dual relationship between intent and extent is called \emph{Galois connection}.} 


The core insight of abstractionism is that concepts are related to each other to form a hierarchical structure, which is formally represented by the dual lattices of intent and extent. 
Although this is clearly an important aspect of concepts, the abstractionist model has also been subject to several criticisms. 
Wittgenstein famously questioned the presumption that concepts can be defined by a set of necessary and sufficient conditions \citep{Wittgenstein1953-nw}. 
With rare exceptions of formal concepts introduced by explicit definitions, most real concepts appear to have a rather loose boundary and defy a clear-cut identification in terms of either intent or extent. 
This is particularly so with biological concepts like species, where variation is the rule rather than exception \citep{Mayr1997-od}. 
Second, traditional Abstractionism has been criticized for being too permissive, allowing for the formation of arbitrary and often unintuitive concepts. For instance, it is often taken to suggest that for any pair of concepts, such as \cpt{cherry} and \cpt{raw meat}, there exists a concept formed by abstracting away their differences, leading to constructs like \cpt{red juicy food}, which many argue lacks coherence and practical relevance (\citealt{Lotze1874-vu}; \citealt{Heis2007-ti}, pp. 95-6).
To prevent such arbitrary abstraction, the abstract operation (i.e. join) must be restricted in some way or another.\footnote{This criticism assumes that the concept lattice is Boolean, an assumption that has tacitly persisted throughout the history of philosophy. However, this criticism does not apply to FCA, which models concepts using complete (but not necessarily Boolean) lattices. In complete lattices, the abstraction operation (join) must take closure, preventing the formation of arbitrary concepts. See \citet[][Chaps. 3 and 7]{Davey2002-ao}} 
In psychology, the classical theory has also been criticized for its failure in capturing  human cognition, in particular \emph{typicality effects}, where object classifications tend to take more time and less reliable with atypical objects (such as eels in the case of fish) than typical ones (such as trout) \citep{Margolis1999-ob}. 
Such effects cannot be explained by the abstractionist theory, which basically sees no differences among items belonging to the same category  as long as they share the same defining attributes. 
Finally in the context of machine learning, creating the knowledge base along the abstractionist line proved to be a challenging task. 
FCA, for instance, requires a table (called ``context'') that lists objects and attributes possessed by them. 
However, identifying what counts as relevant attributes and which objects have those attributes is not a trivial task but often calls for ingenuity informed with domain knowledge. 
Extracting such knowledge from human experts was one of the main obstacles (called ``Feigenbaum's bottleneck'') in early AI research \citep{Gillies1996-jj}. 

These criticisms of abstractionism motivated alternative approaches to model concepts, to which we now turn. 

\section{The Similarity Approach}
As we have seen above, one major criticism against abstractionism was targeted at its essentialism that a concept is definable by a set of necessary and sufficient conditions, or its essence.
Most real concepts seem to lack such essence. 
What, for example, are properties that are shared by all instances of \cpt{game}, which would presumably include football, \emph{Go}, \emph{Tetris}, and so on?
Wittgenstein famously pointed out that what unites these games are not the shared properties but rather \emph{family resemblance}, i.e., they are related via loose similarity relationship among objects. 
Similarity is symmetric but not necessarily transitive: I might resemble my maternal grand father and paternal grand mother, but they do not need to resemble each other. 
Likewise, an arbitrary pair of games, say football and \emph{Tetris}, need not be similar or have shared properties. What make them \cpt{game} is the fact they resemble to other items that together form a cluster of objects we call game. 

%
Similarity judgments depend on properties or standards under considerations.  
Two people may have similar body heights but not weights. 
To measure the similarity between objects, therefore, one first needs to determine a set of relevant properties or dimensions, which can take either continuous, discrete, or binary (yes or no) values, and then plot objects according to their value in each dimension. 
Suppose we have $n$ relevant properties and let $x_i (a)$ be the value of the $i$-th property/dimension of an object $a$. 
One straightforward similarity measurement of a pair $(a,b)$ of objects is a weighted sum of the absolute difference 
\begin{equation}
 \sum_i^n w_i \cdot | x_i(a) - x_i(b)|
\end{equation}
or the Euclidean distance
\begin{equation}
 \sum_i^n \sqrt{w_i \cdot ( x_i(a) - x_i(b))^2},    
\end{equation}
where the weight $w_i$ gives the relative importance of the $i$-the property toward calculating similarity. 
A pair similar in this sense will be mapped closely in the $n$-dimensional property space. 
A concept, then, can be defined as a cluster in the high-dimensional space. 
Such a cluster may be ``cloud-like'' and need not have clear-cut boundaries or threshold values, capturing the idea that concepts are usually not definable by sharp necessary and sufficient conditions. 

The similarity-based view of concepts goes by different names in different fields. 
In philosophy of science, the \emph{Homeostatic Property Cluster} or HPC Theory has argued that property clusters can better account for \emph{natural kinds}, especially those appearing in biological sciences such as biological species, than the definitional approach \citep{Boyd1991-no, Boyd1999-hg}.
The same idea underlies \emph{prototype theory} and \emph{exemplar theory} in cognitive science, which posit that humans classify objects based on their similarity to representative items, referred to as prototypes or exemplar \citep[cf.][ch. 3]{Murphy2004-nk}.
According to prototype theory, an object is classified into a category, say \cpt{cat}, if it is more similar to the prototype of \cpt{cat} than to other prototypes, where a prototype is defined as a  statistical center, such as the mean, derived from past instances of the same kind. 
Exemplar theory, on the other hand, suggests that the standard for classification is set by a representative instance, such as an actual sparrow or robin in the case of birds, rather than a statistical construct. 
These views excel in handling the typicality effects we saw above by explaining the time lag and unreliability of classification judgments as a function of the distance from the prototype or exemplar, and became widely accepted in cognitive science.


The notion of similarity has also played the crucial role in the recent development of machine learning, more specifically in the construction of good \emph{representations}.
Deep neural networks (DNNs) process complex data by representing them as vectors in a latent space, which are then used for the downstream tasks such as image classifications, word predictions, and so on. 
Such representations are expected to be \emph{aligned}, i.e., reflect the actual relationships among objects so that similar things are embedded into a neighboring area \citep{Wang2020-lh}.
For instance, natural language processing (NLP) embeds words in a high-dimensional vector space so that word-vectors with similar meanings cluster together. The similarity between these vectors is measured by the angle formed between them, known as cosine similarity. 
It has been known that this task is accomplished with a simple three-layer network, such as word2vec, and a corpus of modest size \citep{Mikolov2013-xz}.
More generally, the technique called \emph{contrastive learning} is used to learn the similarities and differences between data points in a wide range of tasks \citep{Van_den_Oord2018-jf}. 
It compares given data points with both similar (positive) and dissimilar (negative) examples, learning representations that minimize the distance between positive pairs and maximize the distance between negative pairs, thus capturing the underlying semantic structure of the data. 
Representations learned through contrastive learning have proven highly effective in image classification, NLP, and speech recognition, excelling at distinguishing categories, understanding semantic relationships, and identifying different speakers and phonetic patterns.

Underlying all the above approaches is the mathematical notion of metric space, a space equipped with a metric function that specifies the distance between any two points in the space. 
The absolute distance, Euclidean distance, and cosine similarity are typical examples of metric functions, but there are plenty of other choices.
Moreover, the metric does not need to be defined globally but may differ from point to point, which is the case when the space is not flat throughout but curved (e.g., Riemanian manifolds). 
Determining the metric structure of the conceptual space, therefore, will become the first task of this modeling strategy. 
The next equally important task is to determine the dimensions of the space, which serve as the standards against which the similarity is measured. 
Because two objects can be compared with respect to infinitely many possible aspects, cognitive scientists must specify \emph{a priori} what features count in calculating their similarity. 
In contrast, dimension reduction is an automatic feature of machine learning algorithms. 
In a word embedding algorithm like word2vec, the input dimension is the size of the vocabulary, so if the corpus contains 10000 different word types, each word is a $1 \times 10000$ vector (called one-hot vector). The algorithm is trained to embed these input vectors into the ``latent space'' in the middle layer, whose dimension is typically about a few hundreds. 
The ``features'' or dimensions of the latent space are constructed \emph{a posteriori} through the learning process, and usually lack specific meaning. 

The similarity approach thus highlights the geometrical aspect of concepts, which prove to be crucial in resolving several challenges against abstractionism, such as the typicality effect in cognitive science and learnability from data in machine learning. 
Moreover, it has been reported that word vectors embedded by word2vec allow for certain algebraic operations such as addition and subtractions, so that when one subtracts the word vector \cpt{male} from \cpt{king} and adds \cpt{female}, the resulting vector lie in a close vicinity of that of \cpt{queen} \citep{Mikolov2013-xz}. 
This attests that they are indeed mathematical objects of vectors where addition is already built-in. 

This raises the question of whether the geometrical approach can also accommodate other algebraic operations that are inherently tied to concepts.
Several works have proposed interpreting the negation of a word vector as a form of disambiguation \citep{Widdows2003-cz, Widdows2004-iv, Ishibashi2024-og}. For instance, \cpt{rock NOT band} should correspond to a vector related to geological rocks but unrelated to musical bands. Such an operation can be realized by projection: the word vector \cpt{a NOT b} can be obtained by projecting \cpt{a} onto the subspace orthogonal to the subspace spanned by \cpt{b}. Conversely, disjunction \cpt{a OR b} acts as ambiguation, applying to any item between \cpt{a} and \cpt{b}. This can be represented by the subspace spanned by the vectors \cpt{a} and \cpt{b}.
Widdows observes that with these operations the word vector space has an algebraic structure similar to Quantum logic, whose semantics uses vector spaces instead of sets. 

Another line of research seeks to geometrically represent the hierarchical structure of words, thereby incorporating the abstractionist idea within the similarity framework. 
Poincar\'{e} embedding proposed by \cite{Nickel2017-fy} replaces the standard Euclidean representation space with a hyperbolic space, which is more suitable to encode tree structures. 
Another strategy is to represent words not by points or vectors but by  extended areas like boxes \citep{Vilnis2018-jk, Dasgupta2021-rp}. Then the hierarchical ``is-a'' relationship is easily handled by set inclusion, and the resulting set of regions (i.e., subsets of the entire space) becomes a complete lattice equipped with the disjunction (join) and conjunction (meet) operations \citep{Davey2002-ao}. 
At this point, however, it is unclear whether such attempts will be able to cover other aspects of compositionality that our languages is supposed to have \citep{Margolis1999-ob}.

\section{The Functional Approach}
Above we have seen that the similarity approach was prompted by the criticism on the essentialist aspect of abstractionism. 
The functional approach is inspired by another issue of abstractionism that permits an arbitrary concept formation through the abstraction of random objects, say the concept of \cpt{red juicy food} from cherry and raw meat \citep{Lotze1874-vu, Heis2007-ti}. 
A similar point may be raised against the similarity approach.
Recall that in the similarity approach concepts are represented as regions or ``chunks'' in the metric space. 
Now consider abstracting these concepts. 
The straightforward way to do this is to combine all the regions corresponding to these concepts to form a greater region: for instance, \cpt{mammal} could be formed by combining all such clusters like \cpt{human}, \cpt{dog}, \cpt{whale}, etc. 
But obviously not any amalgamation would do: gerrymandered or isolated patches, like the notorious ``grue'' formed by all green things observed before a certain time $t$ and blue things observed thereafter, should not make a \emph{bona fide} concept \citep{Gardenfors1990-bu}. 
This means that there must be a certain restriction on the shape of a region or cluster for it to count as a concept. 

Such a constraint can be represented by a certain functional relationship. 
According to this idea, a concept is not a mere combination of attributes or items but rather embodies a certain functional relationship among its possible features. 
\cite{Heis2007-ti} attributes this ``functional'' view of concept to Lotze:
\begin{quote}
As a rule, the marks of a concept are not coordinated as all of equal value, but they stand to each other in the most various relative positions, offer to each other different points of attachment, and so mutually determine each other; ... an appropriate symbol for the structure of a concept is not the equation $S = a + b + c + d$, etc, but such an expression as $S = F (a, b, c, \textrm{etc.})$ indicating merely that, in order to give the value of $S, a, b, c,$ etc, must be combined in a manner precisely definable in each particular case, but extremely variable when taken generally. (\citealt{Lotze1874-vu} §28, quoted from \citealt{Heis2007-ti}, p. 283)
\end{quote}
Lotze's point is that a concept cannot be created by combining arbitrary features willy-nilly.
It is rather characterized by its internal functional relationship, through which its possible attributes are constrained by each other. 
For instance, each concept that belong to \cpt{animal} should specify as its attributes the means for locomotion, reproduction, and respiration, etc, like $\cpt{dog} = f(\cpt{walking}, \allowbreak \cpt{viviparous}, \cpt{pulmonary},...)$ \citep{Asano2020-ba}. 
Moreover, these attributes are not independent from each other: e.g., the respiratory system of an organism should match its locomotive means.
Lotze's proposal is that the concept \cpt{animal} should be understood as such an internal relationship that determines possible combinations of animal traits.

The functional relationship embodied by a concept reflects our theoretical knowledge about the objects and world. 
That flying animals usually have lungs instead of gills is a part of our folk biology.\footnote{The French anatomist George Cuvier elevated such functional relationships among anatomical parts into the status of biological law, which he called the \emph{principle of the correlation of parts} \citep{Cuvier1805-re}.}
In this sense, the functional view shares the spirit with the so-called \emph{theory theory} or \emph{knowledge approach} in cognitive science that sees concepts as anchored by our overall understanding of the world \citep[pp. 60-1]{Margolis1999-ob}. 
According to this view, concepts are not mere labels of objects, but rather carry with them significant information about what they are, which is exploited by our reasoning and inference. 
Although there is much room of interpreting what it counts as a ``theory'' behind each concept, one (but obviously by no means only) possibility is to represent it as a functional relationship. 
Take as an example what \cite{Barsalou1985-ln} calls \emph{goal-derived categories}. 
\cpt{healthy food} is one such category, which may include items such as \cpt{kale}, \cpt{whole grains}, and \cpt{lentils}. 
However, its content are not determined merely by a disjunction of these items. 
Rather, we consider a vector-valued function that maps food to its nutritional components such as calories, fiber, proteins, etc., and defines \cpt{healthy food} as an inverse image of certain regions (e.g. a low calorie and high fiber and protein region) in the codomain of such a function. 
Thus the concept \cpt{healthy food} is based on the function that presumably reflects our knowledge in nutritional science. 

Let’s examine how this functional perspective of concepts intertwines with the geometric landscape discussed in the previous section.
As seen in graphs described by equations, a function serves to specify a certain region or hypersurface in the space. 
For instance, $f(x,y) = x^2 + y^2 = 1$ in the two-dimensional Euclid space $\mathbb{R}^2$ defines the origin-centered circle with radius 1. 
In general, a real-valued function $f$ on a space determines a region of the space by its inverse image $f^{-1}(r)$ for $r \in \mathbb{R}$ (which, in the case of the above function, is the circle with radius $r$). 
Moreover, if it is a smooth function $f:M \to \mathbb{R}$ on a manifold $M$, it determines a submanifold $M' \subset M$ as its inverse image $M' = f^{-1}(r)$ for $r \in \mathbb{R}$ under some regularity condition.\footnote{More precisely, the condition is that the inverse image $f^{-1}(r)$ does not contain any \emph{singular points}, where all the partial derivatives of $f$ vanish. See \citet[Sec. 9.3]{Tu2010-zr}.}
Such a submanifold $M'$ represents possible combinations of features under the functional constraint $f$. 
The functional concept, therefore, can be represented as a certain region in the conceptual space, which in cases of smooth regular function constitutes a submanifold.\footnote{However, the opposite does not hold. That is, not every submanifold can be obtained as the inverse image of a regular value of a smooth function.} 

\begin{figure}[h]
    \centering
    \begin{subfigure}[b]{0.45\textwidth}
        \centering
        \includegraphics[width=\textwidth]{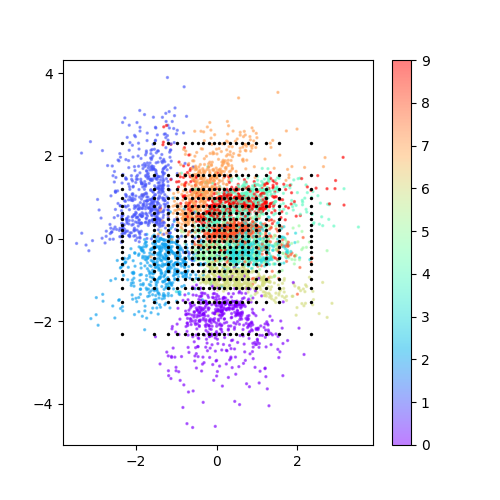}
    \end{subfigure}
    \hfill
    \begin{subfigure}[b]{0.45\textwidth}
        \centering
        \includegraphics[width=\textwidth]{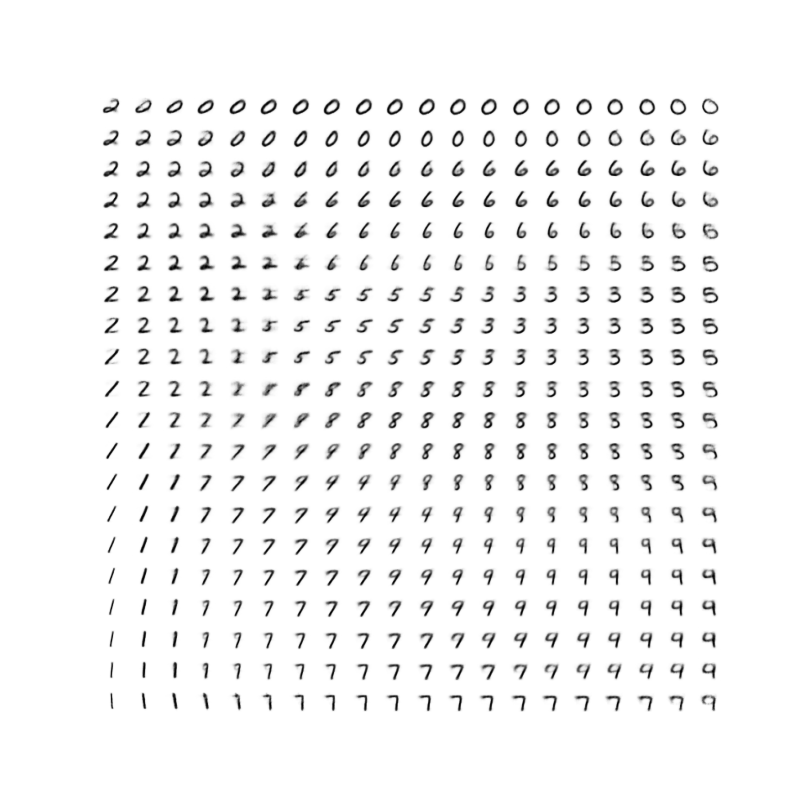}
    \end{subfigure}
    \caption{Manifold learning with VAE. The left plot shows the latent representation space of a VAE model trained on the MNIST dataset, where each digit is represented by a different color, forming (roughly) continuous subregions. As a generative model, the VAE can produce images corresponding to points within this representation space. The right image displays figures generated from the black dots arranged in a grid in the left plot, illustrating the continuous morphing of digit shapes.}
    \label{fig:vae}
\end{figure}

\begin{figure}
    \centering
    \includegraphics[width=0.4\linewidth]{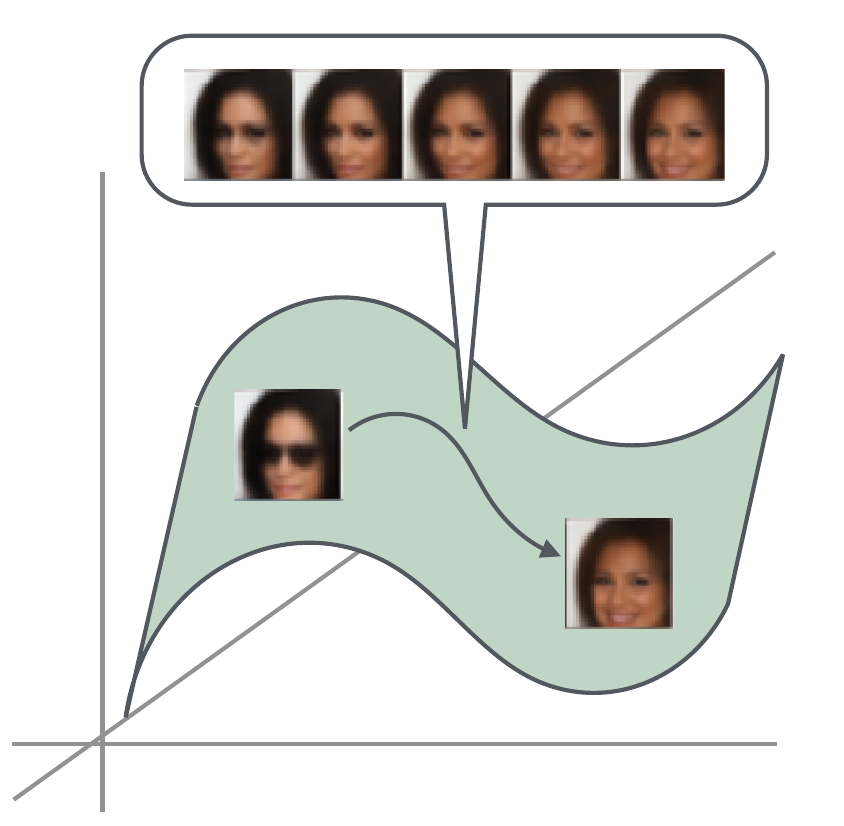}
    \caption{An image of face morphing on a conceptual manifold. If a concept is defined as a submanifold in the representation space, a smooth curve connecting two data points represents a ``morphing'' of one instance (her, a face with glasses) into another (one without). Images are adapted and modified from \citep{Higgins2016-rx}.}
    \label{fig:morphing}
\end{figure}

In the machine learning literature, the \emph{Manifold Hypothesis} postulates that observed data can be represented in low-dimensional regions or submanifold within the high-dimensional representational space, reflecting the fact that not all possible combinations of features are realizable \citep[cf.][Sec. 5.11.3]{Goodfellow2016-xx}.\footnote{But note that the manifold hypothesis does \emph{not} claim that such submanifolds are generally obtained as inverse images of certain functions. See footnote 5.} 
If the hypothesis is correct, different types of objects or concepts should be represented as  different submanifolds in the representational space.
This idea underlies a variety of manifold learning algorithms such as \emph{Variational Auto-Encoder} or VAE \citep{Kingma2013-eo}.
As the image classification algorithm discussed above, VAE embeds objects into the latent representational space, but it adds slight Gaussian noise (variation) in the encoding process. 
The added noise serves to ``smooth'' the representational space, so that objects in the vicinity of the representational space are smoothly connected to each other, making each cluster a manifold-like (Fig. \ref{fig:vae}).
One significant advantage of this algorithm is that one can smoothly ``morph'' one object into another by moving along a curve on the learned manifold. Suppose the algorithm is trained on the dataset containing various human faces. Then each face image is represented by a point in the latent space. Take two such distinct points, and consider moving one of them toward the other. It has been shown that such a motion gives rise to a continuous morphing from one image to the other, say a face with glasses to without \citep[][see also Fig. \ref{fig:morphing}]{Higgins2016-rx}. This suggests that the algorithm represents human faces as a manifold, whose points/faces are connected via a smooth curve.

The possibility of morphing implies that slight displacements in the representation space correspond to possible changes in the world. This, in turn, suggests that the conceptual manifold captures, at least locally around each data point, the actual similarity relationships and constraints among the objects being represented. 
This is not necessarily guaranteed when the similarity space is constructed based on arbitrary sets of attributes. Suppose we decided to map objects, say people, using two features, height and weight. 
But if these properties are functionally related to each other, the real degree of freedom would be close to one, in which case people are actually lying on a one-dimensional curve rather than two-dimensional space. 
Moreover, if the function is nonlinear, the subspace containing data has a non-zero curvature that may vary from points to points. 
In this case, the area or points considered close to a given datapoint by the global metric may actually be far away or indeed inapproachable. 

To capture the real similarity between objects, therefore, it is crucial to consider the functional relationships among the features that constitute the representational space. 
These functional relationships can be viewed as the ``law'' that governs a particular type of object or concept: for instance, the complex functional relationships that presumably underlie the \cpt{face} manifold discussed above would determine possible configurations of facial parts and specify how these components relate to each other. 
The primary premise of the functional approach is that a concept is adequately represented by the law inherent to the type of objects to which the concept applies. 
A proper understanding of a concept/law allows one to not only delineate and anticipate possible variations of a given type, but also predict other features based on some known information. 
In this way, functional concepts highlight aspects of conceptual reasoning that have not been emphasized in either abstractionism or the similarity approach. 

\section{The Invariance Approach}
The generative capability of concept manifolds, which allows one instance to morph into another, highlights a dynamic aspect of concepts, enabling various forms of inductive reasoning about hypothetical changes.
When we classify an object under a certain concept, e.g. identify it as a \cpt{face}, we are at the same time attributing to it a various projections about how its appearance would change or would not change under possible transformations, say in perspective or the passing of time. 
As Cassirer puts it, ``the apprehension of the particular \textit{qua} `existence' involves apprehension of the possibilities of transformation which it contains within itself'' \citep[p. 15]{Cassirer1944-pm}.
In the context of psychological research, the same idea underlies Gibson's pioneering work on \emph{affordance}, which posits that animal perceptions encompass not merely sensory stimuli but also rich information about what the environment offers or affords the organism in terms of possibilities for action. This includes, for example, the organism's anticipation of how visual perception might alter upon moving through its surroundings \citep{Gibson1979-ay}. 
This is echoed by Barsalou's perceptual theory of knowledge, which identifies a concept to be a ``simulator'' that allows the cognitive argent to mentally simulate different aspects and instances of a given category, say chair, in many different circumstances \citep[p. 587]{Barsalou1999-xc}. 


The key insight here is that objects and their appearances change in a systematic fashion, and thus our recognition system must be robust enough to identify the object under different guises, while also being flexible enough to track these changes.
These features are respectively captured by \emph{invariance} or \emph{equivariance} with respect to group actions \citep{Cassirer1944-pm, Hoffman1966-um, Dodwell1983-my, Jantzen2015-gx}.
This framework first models various changes in objects or their appearance as transformations, represented by functions $g: X \to X$ where $X$ is a set of items in question.
For instance, if $X$ is the set of visual images formed on the retina, a shift $g$ in perspective changes one image $x \in X$ to another $g(x) \in X$.
The premise is that the set $G$ of all such transformations forms a \emph{group}, so that 
(i) it contains an identity or ``do nothing'' transformation $e \in G$ such that $e \cdot x = x$ for all $x \in X$; 
(ii) for every transformation $g \in G$ there is an inverse ``cancellation'' $g^{-1} \in G$ such that $g \cdot (g^{-1} \cdot x) = g^{-1} \cdot (g \cdot x) = x$ for all $x \in X$; and 
(iii) transformations are associative, so that $(g_i \cdot g_j) \cdot g_k = g_i \cdot (g_j \cdot g_k)$ for any $g_i, g_j, g_k \in G$.
Changes in perspective, for instance, arguably satisfy these rules and thus present \emph{group actions} on $X$. 

Conceptualization can be understood as a function $\phi:X \to R$ that maps objects or stimuli $x \in X$ to their representations $\phi(x) \in R$.
Such a representation map is called \emph{invariant} with respect to group actions $G$ on $X$ if actions do not affect representation, such that $\phi(x) = \phi(g(x))$ for all $x \in X$ and $g \in G$ (Fig. \ref{fig:invariance} left).
This means that however one transform an object $x$ within the range of $G$, $\phi$ still identifies them as the ``same thing.''
This is arguably a desirable feature of conceptual recognition, as the apprehension of ``what it is''---such as identifying whose face it is---should not depend on a certain range of transformations like perspectival shifts.
In other words, the ability to classify objects or stimuli into one concept, or ``the apprehension of the particular \textit{qua} `existence' '' as Cassirer puts it, involves determining the range of group actions $G' \subset G$ with respect to which representation $\phi$ remains invariant. 

\begin{figure}
    \centering
    \[\begin{tikzcd}
        x & \phi(x) \\
	g(x) & \phi(g (x))
	\arrow["\phi", from=1-1, to=1-2, maps to]
	\arrow["\phi", from=2-1, to=2-2, maps to]
	\arrow["g", from=1-1, to=2-1, maps to]
	\arrow["", from=2-2, to=1-2, equal]
    \end{tikzcd}
    \hspace{1cm}
    \begin{tikzcd}
        x & \phi(x) \\
	g(x) & \phi(g(x)) = \psi(g)(\phi(x))
	\arrow["\phi", from=1-1, to=1-2, maps to]
	\arrow["\phi", from=2-1, to=2-2, maps to]
	\arrow["g", from=1-1, to=2-1, maps to]
	\arrow["\psi(g)", from=1-2, to=2-2, maps to]
    \end{tikzcd}    
    \]
    
    \caption{Invariant (left) and equivariant (right) representation.}
    \label{fig:invariance}
\end{figure}

On the other hand, \emph{equivariance} refers to the aspect of representations that ``tracks'' changes in objects. 
Perspectival shifts, for instance, though leaves invariant the identity of the object, will certainly change the way it appears. 
Such a change $g$ in objects or stimuli should induce a concomitant transformation $\psi(g): R \to R$ of representations, which we also assume to form a group.\footnote{Thus $\psi$ is a group homomorphism from $G$ to the symmetry group (the set of all permutations) of $R$. Note that the invariance is a special case $\psi(g) = e$, where $e$ is the identity transformation of $R$. In this respect, the invariant group actions $G' \subset G$ are the kernel of $\psi$.} 
This group of representation transformations $\psi(G)$ is said to be \emph{equivariant} with respect to $G$ if the right diagram of Fig. \ref{fig:invariance} commutes, i.e., if $\phi(g(x)) = \psi(g)(\phi(x))$ for all $x \in X$ and $g \in G$. 
The left hand side of the equation is the representation of a transformed object $g(x)$, while the right hand side is the the result of applying $\psi(g)$ to the original representation $\phi(x)$. 
If one interprets $\psi(g)$ as a representation of the transformation $g$, the equality demands that actual transformations $G$ are as it were well ``simulated'' in the representation space by $\psi(G)$. 

Invariance and equivariance have been important notions in machine learning, particularly image recognition where a model is tasked to identify an object from images taken in various conditions \citep[cf.][]{Weiler2024-ym}. 
Convolutional neural networks (CNNs) are specifically adopted for such a task: by repeatedly applying the same kind of local pattern detectors over an entire image and pooling that information, CNNs can recognize objects, say a \cpt{face}, regardless of their location within the input image. 
Here the relevant group $G$ consists of all parallel displacements (called translations), so CNNs are said to be \emph{translation invariant}. 
Equivarience, on the other hand, refers to a model's capability to vary outputs in accordance to changes in inputs. For instance, suppose we want our model to not just identify a face in the image but also the direction it is facing. Then our model should be \emph{rotation equivariant}, meaning it can adjust its output to accurately reflect the face’s orientation regardless of how the face is rotated in the image. 

The image morphing we saw in the previous section can be understood in terms of (continuous) group actions on concept manifolds. 
To illustrate this with a simple toy example, consider again a unit circle $S^1$ defined by the equation $f(x, y) = x^2 + y^2 - 1 = 0$.
Take any point $s \in S^1$ on this circle and rotate it by any degree, and it will still belong to the same circle. 
Since the set of all rotations form a group\footnote{A finite rotation is generated by Lie operator $\mathcal{L}_r = -y \frac{\partial}{\partial x} + x \frac{\partial}{\partial y}$. The invariance of the circle under this operator can be understood by the fact that applying this operator to the above function that defines the unit circle, $f(x, y) = x^2 + y^2 - 1$, yields $\mathcal{L}_r f(x, y) = 0$, i.e., it produces no change.}, this means that their action on any point on the circle leaves invariant its identity \emph{as a component of the circle}.
On the other hand, rotation changes the coordinate of a point by sliding it along the circle, so that a point's location on the circle is equivariant. 
Now replace the circle with a more complex manifold of \cpt{face}, and assume that this manifold is represented by a certain functional constraint on the high-dimensional coordinates $\mathbf{z} = (z_1, z_2, \dots, z_n)$ of the latent space such that $f(\mathbf{z}) = 0$. 
Then any concrete solution $\hat{\mathbf{z}}$ of this equation (such that $f(\hat{\mathbf{z}})=0$) corresponds to a particular face image. 
Continuous group actions on this face manifold would map each such solution to another solution, that is, one face to another face, in a smooth fashion. 
In so doing, they transform the specific features of the faces equivariantly, while keeping the identity of being a face invariant.
The characteristics of such continuous group actions capture well what we envision by ``morphing.''

At this point, one might wonder why we have to consider such transformations in the first place. From the invariance approach, the answer is that such a dynamic understanding of objects is an integral part of the concepts we form about them. When we perceive an object and classify it under some concept, say recognizing it as a human \cpt{face}, we are not just labeling a still image. As \cite{Gibson1979-ay} and \cite{Barsalou1999-xc} stress, the conceptualization involves the anticipation of how the appearance of the object will change in accordance with incessant shifts in environmental conditions or movements of the perceiving subject.
For otherwise, our perceptual system cannot integrate successive images into a coherent whole and identify it as \emph{a} thing \citep{Kant1999-lu}. According to the invariance approach, a concept embodies a system of rules or laws according to which the object in question transforms, and such lawful connections are suitably modeled by group actions \citep{Cassirer1944-pm}.
Of course, writing down all the invariance and equivariance properties of a given concept in terms of group actions will be at least a challenging, if not unfeasible, task, except for very general transformations like translations, rotations, dilatations, and so on \citep{Hoffman1966-um, Dodwell1983-my}. Even so, group-theoretic modeling extracts the formal essence of the Gibsonian insight and sheds light on the dynamic---invariant and equivariant---nature of concepts.

Furthermore, the group theoretic consideration prompts a new approach to investigate the dimensionality of concepts. 
As in the case of \cpt{face} morphing, there are multiple ways to alter objects or their perceptions. It is then natural to ask how many distinct transformations exist (namely, the degree of freedom) and what their mutual relationships are, since some of these transformations are interdependent while others are independent (e.g. difference in gender is often, though not necessarily, related to hair style, but not to whether a person is putting eyeglasses or not).
A good conceptualization must handle these issues by distinguishing different types of transformations and encoding the (in)dependence relationships among these transformations, or in other words, it must ``carve nature at its joints.''
Failure to do so results in \emph{entangled representations} that mix up essentially irrelevant features: for instance, if a group action that transforms hair length also affects hair color, the representation is said to be entangled \citep{Higgins2016-rx, Higgins2018-hm}.

\cite{Higgins2018-hm} suggest that representations whose transformations align with the real ``joints'' of nature---namely, \emph{disentangled} representations---can be modeled in terms of the decomposition of group actions. 
The rough idea is as follows. 
Suppose group $G$ acting on the set of representations $Z$ can be written as a product $G = G_1 \times G_2 \times \dots \times G_n$. 
This amounts to the assumption that transformations $G$ are decomposed into distinct types $G_1, G_2, \dots, G_n$. In the case of perspectival changes, for instance, they may be translations, rotations, dilatations, etc. 
Then the representation space $Z$ is said to be disentangled if it is also decomposed in accordance with the group decomposition, that is, there is a decomposition of $Z$ into subspaces $Z = Z_1 \times Z_2 \times \cdots \times Z_n$ such that each $Z_i$ is affected only by $G_i$, namely, for all $g_i \in G_i$ and $z = (z_1, \dots, z_i, \dots, z_n) \in Z$, $g_i(z) =  (z_1, \dots, g_i(z_i), \dots, z_n)$, for $1 \leq i \leq n$.
This definition captures the idea that the representation space $Z$ is properly carved into its subspaces in accordance to the joints determined by decomposed group actions, which in tern represent distinct ways in which the object in question can transform.
When this condition is satisfied, we can interpret each dimension $Z_i$ of $Z$ as an independent feature of an object, say \cpt{hair color} or \cpt{gender}.

In the similarity approach discussed in Section 3, features or attributes of objects defined the coordinates against which the similarity between the objects is calculated. 
In Section 4, we saw that what we perceive as independent features may be functionally related to each other, so that the real degrees of freedom are smaller than appearance.
Group-theoretic considerations shed light on this fact from a different angle, suggesting that the degrees of freedom reflects the dynamic nature of objects under consideration and are determined by possible ways in which these objects vary. 

\section{Discussion}
The preceding sections have examined various theories on concepts and representations in philosophy, cognitive science, and machine learning, classifying them into four categories: abstractionist, similarity, functional, and invariance approaches. This section compares these views from a meta-perspective, exploring their connections and deriving implications for further studies. By identifying the intersections and differences among these approaches, this discussion section aims to provide a clearer framework for interdisciplinary studies of concepts and representations.

The first axis for comparison is the role of concepts. Arguably, concepts play various roles in classification, learning, communication, problem solving, and so on; but these roles can be further understood through the lens of two major functionalities: descriptive and inferential.
In the descriptive use, concepts serve to characterize and summarize data in various formats. Among the approaches discussed so far, the abstractionist and similarity approaches are particularly motivated by these descriptive tasks. 
The major goal of abstractionists is to classify objects in the hierarchical manner, by constructing a conceptual lattice from a data table or ``context'' that lists items and their observed properties (Section 2). 
Given a similar (but possibly continuous) dataset, the similarity approach visualizes mutual relationships between items in a two or more dimensional space based on their attributes and delineates concepts as clustered points that are close to each other (Section 3).
These outcomes---lattices or plots---reveal the inherent structure of data not visible in the original format.

Concepts also play a central role in inferential tasks, e.g., for the purpose of predicting future events or possible behaviors of objects. 
By identifying an object in front of me as a \cpt{dog}, I can anticipate what it will and will not do: it may bark, chase after balls, and sniff around, but probably not climb trees or purr like a cat. Though very simplistic, it is nonetheless a \emph{bona fide} act of prediction: based on some observable cues like floppy ears, I derive its behavioral characteristics that I have not yet seen, at least with respect to this particular dog in front of me. 
Such inferential role of concepts is the primary focus of the functional perspective (Section 4). Inferences from one set of attributes to another exploit the information about internal constraints on the possible combinations of attributes. This information, encoded in the functional form, constitutes our understanding of concepts. 
Alternatively, the invariance approach captures inference from a different perspective, by identifying a concept as a set of transformation rules that govern possible changes in the object (Section 5). Such equivariant group actions highlight the inferential role of concepts that allows us to simulate how objects or their perceptions transform in accordance with changes in the environment or the perceiving subject.

The above discussion should not imply that models are inherently connected to or designed for particular roles.  
Indeed, there is no contradiction in taking a conceptual lattice as representing internal constraints of concepts or using the similarity space for the purpose of prediction, as is quite common in natural language processing. 
The descriptive/inferential contrast rather concerns the modeling purpose: namely whether one is interested in data themselves, or in some underlying structure of ``uniformity of nature'' from which data are sampled. 
In the former context, mathematical models are taken as a sort of descriptive statistics that serve for the economy of thought; while in the latter, they are interpreted as representing a generative mechanism that lies beyond any particular data \citep{Otsuka2022-eq}. 
These two desiderata are often in conflict: the more one tries to adjusts one's concepts as faithful as possible to given data, the more likely they are to overfit the data, resulting in a loss of  predictive ability \citep{Forster1994-vj}.
A good descriptive model needs not be a good inferential model, and vice versa. 
Therefore, when evaluating a particular model, it is essential to clearly specify the criteria by which it is being evaluated.

Our second point of consideration concerns the relationship between concepts and their defining features.
In the philosophical and cognitive science literature, concepts have been defined or characterized in terms of familiar and explicit features we use in everyday life, such as color, size, shape, and so on. Accordingly, conceptual hierarchies or similarity maps have been build and evaluated on the basis of a pre-specified set of mostly ostensible features. This represents a ``feature-first'' approach, where features act as the raw material for thoughts, with concepts being formed by combining or mixing these features.
In contrast, features in representation learning are not fixed beforehand, but rather are extracted from data as axes or dimensions that constitute the latent space. In addition, the meaning of these extracted features is not given \emph{a priori}, but must be determined by unpacking the internal intricate structure of the trained neural network through detailed analysis. This can be seen as a ``representation-first'' approach, considering representations to be epistemologically prior to features.

These two approach faces different challenges. 
The main challenge of the feature-first approach lies in its empirical adequacy.
A common criticism against abstractionism or the classic view of concepts points out that it is simply impossible to define concepts, such as \cpt{game} or \cpt{human}, through combinations of existing features or traits \citep{Wittgenstein1953-nw, Boyd1991-no}. 
Moreover, it is not clear which features should be considered to define concepts. Outcomes of formal concept analysis heavily depend on the list of attributes used to characterize data (Section 2). And Feigenbaum's bottleneck highlights the difficulty of determining the relevant attributes for the problem at hand.
A similar issue arises for prototype and exemplar theorists in determining the appropriate attributes/dimensions and their relative importance in constructing the similarity space, which significantly affects similarity judgments \citep{Murphy2004-nk}.

With its successful applications to a range of empirical problems, the representation-first approach seems to be overcoming all the above issues:  deep learning models are able to learn robust representations (with some reservations discussed shortly) that automatically extract relevant features from data, capture complex patterns, and generalize well to new situations. 
The problem, however, is that these features generally resist intuitive interpretation. One thus needs to read off meanings from trained models, but to do so requires a clear understanding of what one is looking for---that is, what are meaningful features?
The idea of disentangled representation discussed in Section 5 is one attempt to explicate what we consider meaningful features of a representation in terms of independent group actions. 
Understanding features is crucial not just for interpretability but also to enhance model performance, ensure robustness, and improve fairness in machine learning applications \citep{Lipton2016-ir}. 
It is pointed out that the well-known phenomenon of \emph{adversarial attack}, where deep learning models misclassify objects due to the addition of small, often imperceptible perturbations to the input data, is a consequence of the model's using complex, high-dimensional features that are not necessarily aligned with human-perceptible features \citep{Ilyas2019-cc}.
Also, understanding features allows data scientists to identify and mitigate biases in machine learning models by examining how different features influence predictions \citep{Ribeiro2016-dt}.
These efforts can be understood as attempts to identify the concepts (representations) used by machines and analyze them into components (features) amenable to human understanding. 

The feature-first and representation-first approaches can thus be understood as akin to digging a tunnel from opposite sides.
The former starts with a set of explicit features and builds complex concepts in a bottom-up way, while the latter aims to break down given representations into understandable pieces in a top-down fashion.
The important challenge in contemporary concept research is to make these approaches meet halfway.

The final, but not least, point concerns the relationship among the four approaches discussed in this paper. 
In his influential book, Edouard \cite{Machery2009-nw} argued for the disunity of concepts, challenging the traditional view that concepts are a unified phenomenon within cognitive science. His view is that what are labeled ``concepts'' actually involve heterogeneous mental kinds with different functionalities, purposes, and empirical bases. 
Be that as it may, mathematical considerations naturally suggest the \emph{logical} relationships between different conceptual models. 
Indeed, we have already seen some of such attempts in the present paper. 
The group-theoretic analysis of disentangled representations can be thought as an attempt to integrate the theoretic aspect of concepts (encoded by group operations) and their similarity-based aspect (represented by a manifold). 
In Section 3, we have seen some recent works in natural language processing that aim to encode the hierarchical structure of concepts into the word vector space by using non-Euclidean (hyperbolic) spaces \citep{Nickel2017-fy} or representing words by boxes instead of vectors \citep{Vilnis2018-jk}. If successful, this line of research will reconcile the abstractionist and similarity approaches, which have been considered rivals in both philosophy and cognitive science literature.

Underlying these studies is the overarching theme of the relationship between geometry and algebra.
Lattices and groups are algebraic in nature, while metric spaces and manifolds have a clear geometric character. Hence the four approaches discussed in this paper can each be seen as shedding light on the geometric or algebraic aspects of concepts.
\cite{Kant1999-lu} was the first to make a clear distinction between, and propose a unification of, these two aspects of human cognition---namely \emph{sensibility} equipped with a  geometric form, and \emph{understanding} that follows logical and algebraic principles. 
Over two centuries after Kant, the contemporary machine learning research is trying to integrate the both components to understand and improve the performance of neural networks, with the aid of much more advanced mathematical machinery than those available to Kant, including non-Euclidean geometry, topology, and group or gauge theory \citep{Sanborn2024-tc}.
Just as Kant was inspired by Newtonian physics in his time, these developments in machine learning will provide new insights into the philosophical understanding of concepts.


\section{Conclusion}
This paper has explored the connections among various approaches to understanding concepts in philosophy, cognitive science, and machine learning, with a particular focus on their mathematical nature. By categorizing these approaches into Abstractionism, the Similarity Approach, the Functional Approach, and the Invariance Approach, we have highlighted the distinct yet interconnected ways in which concepts are represented, organized, and learned across different disciplines.

Each approach offers unique insights into the nature of concepts. Abstractionism provides a structured, hierarchical framework that has influenced both philosophy and early AI research. The Similarity Approach, with its focus on resemblance and metric spaces, has been instrumental in both psychological theories and modern machine learning techniques, such as word embeddings. The Functional Approach introduces a dynamic perspective, emphasizing the internal relationships and constraints that govern concept formation, which aligns with the manifold learning methods used in generative models. Finally, the Invariance Approach, rooted in group theory, underscores the importance of understanding how concepts remain stable under transformations, a principle that is central to the robustness and generalizability of deep learning models.

This paper has also underscored the importance of interdisciplinary exchange. Philosophical insights into concepts can guide the development and refinement of computational models, while empirical findings in cognitive science and machine learning can offer new perspectives on longstanding philosophical questions. As AI continues to advance, the need for a deeper understanding of the representations used by machines becomes increasingly critical. By integrating the mathematical rigor of machine learning with the conceptual analysis of philosophy, we can move towards more transparent, interpretable, and robust models.

Needless to say, the exploration presented in this paper is only a preliminary sketch of the vast landscape of concept representation across disciplines. Notably, the implications of advanced machine learning models such as Attention mechanisms and Diffusion models have not been fully addressed, leaving significant avenues for further inquiry. Additionally, the relationships among the four categories---Abstractionism, the Similarity Approach, the Functional Approach, and the Invariance Approach---require deeper examination, as there may be further connections and insights to uncover. Consequently, systematic and comprehensive investigation remains necessary to deepen our philosophical and mathematical understanding of these interrelated approaches to concepts.

\section{Acknowledgment}
This work was first presented at the Philosophy of Contemporary and Future Science Seminar Series at Lingnan University and was later uploaded to YouTube\footnote{\url{https://www.youtube.com/watch?v=Lqt7TgYk8rU}}. I am grateful for the insightful feedback on these occasions. I also thank Han Bao, Ryosuke Igarashi, and Sho Yokoi for their valuable feedback. This work is supported by the JSPS Grant-in-Aid (KAKENHI) 23K25259.

\bibliographystyle{apalike}
\bibliography{references}

\begin{thebibliography}{}

\bibitem[Asano, 2020]{Asano2020-ba}
Asano, M. (2020).
\newblock Lotze's theory of concepts in logik (1874).
\newblock {\em Historia Philosophiae}, 62:103–128.

\bibitem[Barsalou, 1985]{Barsalou1985-ln}
Barsalou, L.~W. (1985).
\newblock Ideals, central tendency, and frequency of instantiation as
  determinants of graded structure in categories.
\newblock {\em J. Exp. Psychol. Learn. Mem. Cogn.}, 11(4):629--654.

\bibitem[Barsalou, 1999]{Barsalou1999-xc}
Barsalou, L.~W. (1999).
\newblock Perceptual symbol systems.
\newblock {\em Behav. Brain Sci.}, 22(4):577--609; discussion 610--60.

\bibitem[Boyd, 1999]{Boyd1999-hg}
Boyd, R. (1999).
\newblock Homeostasis, species, and higher taxa.
\newblock In Wilson, R.~A., editor, {\em Species: New Interdisciplinary
  Essays}, page 141–186. MIT Press, Cambridge, MT.

\bibitem[Boyd, 1991]{Boyd1991-no}
Boyd, R.~N. (1991).
\newblock {Realism, anti-foundationalism and the enthusiasm for natural kinds}.
\newblock {\em Philos. Stud.}, 61(1-2):127 148.

\bibitem[Buckner, 2019]{Buckner2019-it}
Buckner, C. (2019).
\newblock {Deep learning: A philosophical introduction}.
\newblock {\em Philosophy Compass}, 14(10).

\bibitem[Buckner, 2023]{Buckner2023-wx}
Buckner, C. (2023).
\newblock Transformational abstraction in deep neural networks.
\newblock In {Craver, C Klein, C Haugeland,}, editor, {\em Mind Design III}.
  MIT Press.

\bibitem[Cassirer, 1944]{Cassirer1944-pm}
Cassirer, E. (1944).
\newblock {The concept of group and the theory of perception}.
\newblock {\em Philosophy and Phenomenological Research}, 5(1):1--36.

\bibitem[Cohen and Welling, 2016]{Cohen2016-yq}
Cohen, T.~S. and Welling, M. (2016).
\newblock Group equivariant convolutional networks.
\newblock {\em arXiv [cs.LG]}.

\bibitem[Cuvier, 1805]{Cuvier1805-re}
Cuvier, G. (1805).
\newblock {\em Leçons d'anatomie comparée}.
\newblock Paris: Baudouin.

\bibitem[Dasgupta et~al., 2021]{Dasgupta2021-rp}
Dasgupta, S.~S., Boratko, M., Mishra, S., Atmakuri, S., Patel, D., Li, X.~L.,
  and McCallum, A. (2021).
\newblock {Word2Box}: Capturing set-theoretic semantics of words using box
  embeddings.
\newblock {\em arXiv [cs.CL]}.

\bibitem[Davey and Priestley, 2002]{Davey2002-ao}
Davey, B.~A. and Priestley, H.~A. (2002).
\newblock {\em Introduction to lattices and Order}.
\newblock Cambridge University Press, Cambridge, England.

\bibitem[Dodwell, 1983]{Dodwell1983-my}
Dodwell, P.~C. (1983).
\newblock The lie transformation group model of visual perception.
\newblock {\em Perception \& Psychophysics}, 34(1):1--16.

\bibitem[Forster and Sober, 1994]{Forster1994-vj}
Forster, M. and Sober, E. (1994).
\newblock {How to Tell when Simpler, More Unified, or Less Ad Hoc Theories will
  Provide More Accurate Predictions}.
\newblock {\em Br. J. Philos. Sci.}, 45(1):1 35.

\bibitem[Gibson, 1979]{Gibson1979-ay}
Gibson, J.~J. (1979).
\newblock {\em The Ecological Approach to Visual Perception}.
\newblock Houghton Mifflin Company, Boston, MA.

\bibitem[Gillies, 1996]{Gillies1996-jj}
Gillies, D. (1996).
\newblock {\em Artificial Intelligence and Scientific Method}.
\newblock Oxford University Press, Oxford and New York.

\bibitem[Goodfellow et~al., 2016]{Goodfellow2016-xx}
Goodfellow, I., Bengio, Y., and Courville, A. (2016).
\newblock {\em {Deep Learning}}.
\newblock The MIT Press, Cambridge, MA.

\bibitem[Gärdenfors, 1990]{Gardenfors1990-bu}
Gärdenfors, P. (1990).
\newblock Induction, conceptual spaces and {AI}.
\newblock {\em Philos. Sci.}, 57(1):78--95.

\bibitem[Heis, 2007]{Heis2007-ti}
Heis, J. (2007).
\newblock {\em The fact of modern mathematics: Geometry, logic, and concept
  formation in Kant and Cassirer}.
\newblock PhD thesis, University of Pittsburgh.

\bibitem[Higgins et~al., 2018]{Higgins2018-hm}
Higgins, I., Amos, D., Pfau, D., Racaniere, S., Matthey, L., Rezende, D., and
  Lerchner, A. (2018).
\newblock Towards a definition of disentangled representations.
\newblock {\em arXiv [cs.LG]}.

\bibitem[Higgins et~al., 2016]{Higgins2016-rx}
Higgins, I., Matthey, L., Pal, A., Burgess, C., Glorot, X., Botvinick, M.,
  Mohamed, S., and Lerchner, A. (2016).
\newblock beta-{VAE}: Learning basic visual concepts with a constrained
  variational framework.

\bibitem[Hoffman, 1966]{Hoffman1966-um}
Hoffman, C. (1966).
\newblock The lie algebra of visual perception.
\newblock {\em Journal of Mathematical Psychology}, 3:65--98.

\bibitem[Igarashi, 2023]{Igarashi2023-zh}
Igarashi, R. (2023).
\newblock An essay on the history of philosophy of information: Port-royal
  logic, leibniz, and kant.
\newblock {\em The Journal of Philosophical Studies (The Tetsugaku Kenkyu)},
  609:84--104.

\bibitem[Ilyas et~al., 2019]{Ilyas2019-cc}
Ilyas, A., Santurkar, S., Tsipras, D., Engstrom, L., Tran, B., and Madry, A.
  (2019).
\newblock Adversarial examples are not bugs, they are features.
\newblock {\em Adv. Neural Inf. Process. Syst.}, pages 125--136.

\bibitem[Ishibashi et~al., 2024]{Ishibashi2024-og}
Ishibashi, Y., Yokoi, S., Sudoh, K., and Nakamura, S. (2024).
\newblock Subspace representations for soft set operations and sentence
  similarities.
\newblock In Duh, K., Gomez, H., and Bethard, S., editors, {\em Proceedings of
  the 2024 Conference of the North American Chapter of the Association for
  Computational Linguistics: Human Language Technologies (Volume 1: Long
  Papers)}, pages 3512--3524, Mexico City, Mexico. Association for
  Computational Linguistics.

\bibitem[Jantzen, 2015]{Jantzen2015-gx}
Jantzen, B.~C. (2015).
\newblock {Projection, symmetry, and natural kinds}.
\newblock {\em Synthese}, 192(11):3617--3646.

\bibitem[Kant, 1999]{Kant1999-lu}
Kant, I. (1999).
\newblock {\em Critique of Pure Reason}.
\newblock Cambridge University Press.

\bibitem[Kingma and Welling, 2013]{Kingma2013-eo}
Kingma, D.~P. and Welling, M. (2013).
\newblock Auto-encoding variational bayes.
\newblock {\em arXiv [stat.ML]}.

\bibitem[Lipton, 2016]{Lipton2016-ir}
Lipton, Z.~C. (2016).
\newblock The mythos of model interpretability.
\newblock {\em arXiv [cs.LG]}.

\bibitem[Lotze, 1874]{Lotze1874-vu}
Lotze, H. (1874).
\newblock {\em Logik. Drei Bücher vom Denken, vom Untersuchen und vom
  Erkennen}.
\newblock S. Hirzel.

\bibitem[Lucas and Van Der~Gaag, 1991]{Lucas1991-gv}
Lucas, P. and Van Der~Gaag, L. (1991).
\newblock {\em Principles of Expert Systems}.
\newblock Addison-Wesley Publishing Company.

\bibitem[Machery, 2009]{Machery2009-nw}
Machery, E. (2009).
\newblock {\em Doing Without Concepts}.
\newblock Oxford University Press, USA.

\bibitem[Margolis and Laurence, 1999]{Margolis1999-ob}
Margolis, E. and Laurence, S., editors (1999).
\newblock {\em Concepts: Core readings}.
\newblock The MIT Press, Cambridge, MA.

\bibitem[Mayr, 1997]{Mayr1997-od}
Mayr, E. (1997).
\newblock {\em Evolution and the diversity of life: Selected essays}, volume
  1997.
\newblock Harvard University Press.

\bibitem[McCarthy and Hayes, 1969]{McCarthy1969-ng}
McCarthy, J. and Hayes, P.~J. (1969).
\newblock Some philosophical problems from the standpoint of artificial
  intelligence.
\newblock In Meltzer, B. and Michie, D., editors, {\em Machine Intelligence},
  volume~4, pages 463--502. Edinburgh University Press.

\bibitem[Mikolov et~al., 2013]{Mikolov2013-xz}
Mikolov, T., Chen, K., Corrado, G., and Dean, J. (2013).
\newblock Efficient estimation of word representations in vector space.
\newblock {\em arXiv [cs.CL]}.

\bibitem[Murphy, 2004]{Murphy2004-nk}
Murphy, G. (2004).
\newblock {\em The Big Book of Concepts}.
\newblock The MIT Press.

\bibitem[Nickel and Kiela, 2017]{Nickel2017-fy}
Nickel, M. and Kiela, D. (2017).
\newblock Poincaré embeddings for learning hierarchical representations.
\newblock {\em Adv. Neural Inf. Process. Syst.}, abs/1705.08039.

\bibitem[Otsuka, 2022]{Otsuka2022-eq}
Otsuka, J. (2022).
\newblock {\em Thinking About Statistics: The Philosophical Foundations}.
\newblock Routledge.

\bibitem[Ribeiro et~al., 2016]{Ribeiro2016-dt}
Ribeiro, M.~T., Singh, S., and Guestrin, C. (2016).
\newblock {Why Should I Trust You?: Explaining the Predictions of Any
  Classifier}.
\newblock {\em KDD '16: Proceedings of the 22nd ACM SIGKDD International
  Conference on Knowledge Discovery and Data Mining}, pages 1135--1144.

\bibitem[Sanborn et~al., 2024]{Sanborn2024-tc}
Sanborn, S., Mathe, J., Papillon, M., Buracas, D., Lillemark, H.~J., Shewmake,
  C., Bertics, A., Pennec, X., and Miolane, N. (2024).
\newblock Beyond euclid: An illustrated guide to modern machine learning with
  geometric, topological, and algebraic structures.
\newblock {\em arXiv [cs.LG]}.

\bibitem[Sowa, 2000]{Sowa2000-ps}
Sowa, J.~F. (2000).
\newblock {\em Knowledge Representation: Logical, Philosophical, and
  Computational Foundations}.
\newblock Brooks/Cole.

\bibitem[Tu, 2010]{Tu2010-zr}
Tu, L.~W. (2010).
\newblock {\em An Introduction to Manifolds}.
\newblock Springer New York.

\bibitem[van~den Oord et~al., 2018]{Van_den_Oord2018-jf}
van~den Oord, A., Li, Y., and Vinyals, O. (2018).
\newblock Representation learning with contrastive predictive coding.
\newblock {\em arXiv [cs.LG]}.

\bibitem[Vilnis et~al., 2018]{Vilnis2018-jk}
Vilnis, L., Li, X., Murty, S., and McCallum, A. (2018).
\newblock Probabilistic embedding of knowledge graphs with box lattice
  measures.
\newblock {\em arXiv [stat.ML]}.

\bibitem[Wang and Isola, 2020]{Wang2020-lh}
Wang, T. and Isola, P. (2020).
\newblock Understanding contrastive representation learning through alignment
  and uniformity on the hypersphere.
\newblock In Iii, H.~D. and Singh, A., editors, {\em Proceedings of the 37th
  International Conference on Machine Learning}, volume 119 of {\em Proceedings
  of Machine Learning Research}, pages 9929--9939. PMLR.

\bibitem[Wang et~al., 2024]{Wang2024-wy}
Wang, X., Chen, H., Tang, S., Wu, Z., and Zhu, W. (2024).
\newblock Disentangled representation learning.
\newblock {\em IEEE Trans. Pattern Anal. Mach. Intell.}, PP.

\bibitem[Weiler, 2024]{Weiler2024-ym}
Weiler, M. (2024).
\newblock {\em Equivariant and Coordinate Independent Convolutional Networks: A
  Gauge Field Theory of Neural Networks}.
\newblock PhD thesis.

\bibitem[Widdows, 2004]{Widdows2004-iv}
Widdows, D. (2004).
\newblock {\em Geometry and meaning}.
\newblock CSLI Publications, Stanford, CA.

\bibitem[Widdows and Peters, 2003]{Widdows2003-cz}
Widdows, D. and Peters, S. (2003).
\newblock Word vectors and quantum logic: Experiments with negation and
  disjunction.
\newblock {\em Mathematics of language}, 8:141--154.

\bibitem[Wittgenstein, 1953]{Wittgenstein1953-nw}
Wittgenstein, L. (1953).
\newblock {\em Philosophical Investigations}.
\newblock Macmillan Publishing Company.

\end{thebibliography}

\end{document}